\documentclass[conference]{IEEEtran}

\usepackage{balance}
\IEEEoverridecommandlockouts 
\usepackage[]{graphicx}
\usepackage{algorithmic}
\usepackage{algorithm}
\usepackage{listings}
\usepackage[OT4,T1]{fontenc}
\usepackage[cmex10]{amsmath}
\usepackage{url}
\usepackage{multirow}

\usepackage{placeins}  

\title{Tag and correct: high precision post-editing approach to correction of speech recognition errors}

\author{
    \IEEEauthorblockN{Tomasz Ziętkiewicz}
    \IEEEauthorblockA{
        Adam Mickiewicz University in Pozna\'{n}\\
        ul.\ Uniwersytetu Pozna\'{n}skiego 4, 61-614 Pozna\'{n}, Poland\\
        Samsung R\&D Institute Poland \\
        Plac Europejski 1, 00-844 Warsaw, Poland \\
        Email: t.zietkiewic@samsung.com, tomasz.zietkiewicz@amu.edu.pl}
}

\begin{document}
\maketitle
\begin{abstract}
This paper presents a new approach to the problem of correcting speech recognition errors by means of post-editing. It consists of using a neural sequence tagger that learns how to correct an ASR (Automatic Speech Recognition) hypothesis word by word and a corrector module that applies corrections returned by the tagger.
The proposed solution is applicable to any ASR system, regardless of its architecture, and provides high-precision control over errors being corrected. This is especially crucial in production environments, where avoiding the introduction of new mistakes by the error correction model may be more important than the net gain in overall results.
The results show that the performance of the proposed error correction models is comparable with previous approaches while requiring much smaller resources to train, which makes it suitable for industrial applications, where both inference latency and training times are critical factors that limit the use of other techniques. 

\end{abstract}
\noindent\textbf{Index Terms}: speech recognition, error correction, post-processing, post-editing, natural language processing

\section{Introduction}
\IEEEPARstart{A}utomatic Speech Recognition (ASR) models have been developed for over 70 years. During this time, they evolved from machines that could recognize single digits spoken by one person, created for demonstration purposes without production use back in the 1950s \cite{bell_digits}, to omnipresent voice assistants and speech transcription engines used every day by millions of people around the world. Although speech recognition technology has reached maturity and is production-ready, it is still far from perfect. Professional human transcribers do not reach 100\% transcription accuracy, and although recent deep neural network-powered speech recognition systems are reported to slightly outperform humans \cite{Xiong2016AchievingHP}, they still make mistakes. Furthermore, the near-perfect results of Word Error Rate (WER) below $2\%$ are often reported on popular benchmark datasets such as the test subset of the Librispeech corpus \cite{Librispeech}. When evaluated on corpora from different domains or recorded under different conditions, the results are far from perfect and require adaptation \cite{narayanan2018}. In real-world production settings, input to a speech recognition system may change frequently, caused by changes in topics that are interesting to users. Changes in the real world, such as the COVID-19 pandemic, influence the vocabulary used by speakers and may lead to out-of-vocabulary errors.
The adaptation process of ASR models takes considerable time and resources. In some settings, direct adaptation of the model is impossible, for example, when using a cloud-based speech recognition service. One way to address the imperfections of speech recognition models mentioned above is to improve their results in a post-editor, a module operating on the textual output of a speech recognition model. 
Typically, in production environments, the post-editor provides a means of correcting ASR errors manually, for example, with hand-written regular expressions. The process of creating and maintaining such corrections is laborious and requires a lot of experience \cite{jeziorski-EtAl:2020:LREC2020IndustryTrack}.
Therefore, the post-editor can include an error correction model which learns how to correct errors of a particular ASR system. This approach can be applied regardless of an ASR model architecture, also for systems where direct modification of the model is impossible. It requires only text data to train, and its training requires considerably less resources and time than performing the adaptation of a speech recognition model. This paper presents such an error correction model. In Section \ref{sec:related} we present an overview of previous work on the subject. Section \ref{sec:data} describes the data used for training and evaluation. Details of the proposed approach are presented in Section \ref{sec:method}. Results and conclusions can be found in Sections \ref{sec:results} and \ref{sec:conclusions}. 

\section{Related work} \label{sec:related}
For a review of ASR error detection and correction systems together with a description of ASR evaluation metrics see \cite{ASREDC:review}.
Cucu et al. \cite{cucu_statistical_2013} propose error correction using SMT (Statistical Machine Translation) model trained on a relatively small parallel corpus of 2000 ASR transcripts and
their manually corrected versions. At an evaluation time, the model is used to ``translate'' ASR
hypothesis into corrected form. The system achieves $10.5 \%$ relative WER improvement by reducing WER of the baseline ASR system from $11.4$ to $10.2$.
A similar approach, but using a neural LSTM sequence-to-sequence model and trained on a much larger dataset (40M utterances), is presented in \cite{guo_spelling_2019}. To produce ASR hypotheses, the authors use a speech corpus generated from plain text data with a text-to-speech (TTS) system. In addition to the spelling correction model, authors experiment with
improving the results of an end-to-end ASR system by incorporating an external language model and a combination of the two approaches. The proposed system achieves satisfactory
results ($19\%$ relative WER improvement and $29\%$ relative WER improvement with additional LM re-scoring, with baseline ASR WER of $6.03$) but requires a large speech corpus or high-quality TTS system to generate such corpus from a plain text.
One of the recent works \cite{leng2021fastcorrect} presents an error correction model for Mandarin. The authors stress the importance of a low latency of ASR error correction model A realroduction environments and propose a non-autoregressive transformer model, faster than its autoregressive counterparts. The model is modeled on a large, artificially created parallel corpus of correct-incorrect sentence pairs, generated by randomly deleting, inserting, and replacing words in a text corpus. Real ASR corrections dataset is used to fine-tune the model to a specific ASR system. Relative WER reduction reported by authors on a publicly available testset is $ 13.87 $, which is slightly worse than an autoregressive model ($ 15.53 $) while introducing over 6 times lower latency ($21ms$).

A similar approach to the one presented in this work is proposed in \cite{malmi-etal-2019-encode}, but serves different tasks (grammatical error correction), operates on a poorer set of edit operations and uses different tagging models.

\section{Data} \label{sec:data}
\begin{figure}[htbp]
  \centering
    \includegraphics[scale=0.6]{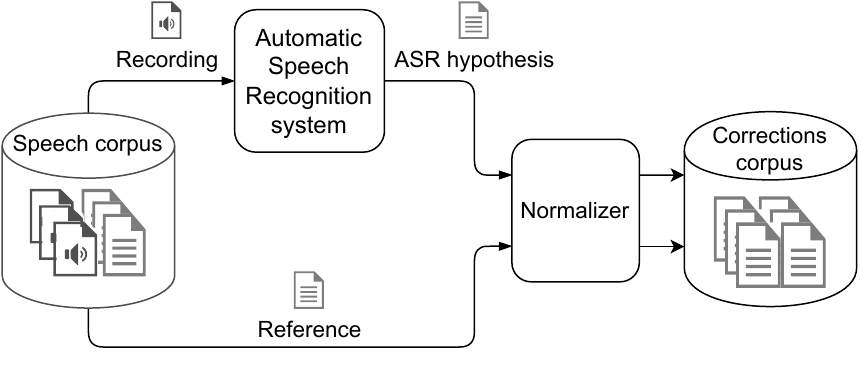}
    \caption{Data preparation pipeline}
    \label{fig:data-pipeline}
\end{figure}

We performed experiments for 3 European languages: Spanish, French, and German. 
The presented error correction models were trained and evaluated on pairs of ASR hypotheses and corresponding reference sentences. To create a corpus of such pairs, recordings from speech corpora for each language were processed using a corresponding model of an end-to-end speech recognition system. Reference transcriptions from speech corpora were paired with their corresponding hypotheses, creating parallel corpora of corrections for each language.
To discard any differences caused by different normalization of transcriptions in speech corpora and in ASR output, we additionally performed 
automatic normalization of both reference and hypotheses sentences by lowercasing, removing punctuation characters and inverse-normalizing numbers. The data preparation pipeline is presented on Figure \ref{fig:data-pipeline}.
For an example of a freely available corpus of ASR corrections for Polish 
prepared with the same pipeline, see \cite{Kubis19b}.

Speech corpora, used to create corrections corpora, contain utterances used for virtual assistant development. They include commands and questions targeted at the assistant. Some of them are recorded in a studio for development purposes, and some originate from usage data, after performing anonymization.

Data statistics are shown in Table \ref{tab:data}.
Data sets were randomly divided into training, development, and test subsets in a proportion 8:1:1.

	\begin{table}
	    \caption{Datasets statistics }
        \label{tab:data}
        \centering
		\begin{tabular}{|l|r|r|r|r|r|}
    		\hline
    		&  \textbf{de-DE} & \textbf{es-ES} & \textbf{fr-FR}  \\
    		\hline
    		Sentences  & $12 242$ & $16 905$ & $7180$ \\
    		Tokens & $37 955$ & $55 567$ & $28 004$  \\
    		\hline
		\end{tabular}
		\end{table}

For a description of tagger training data preparation process, see section \ref{sec:tagging_data}.

\section{Method} \label{sec:method}
\begin{figure}[htbp]
   \centering
     \includegraphics[scale=1.0]{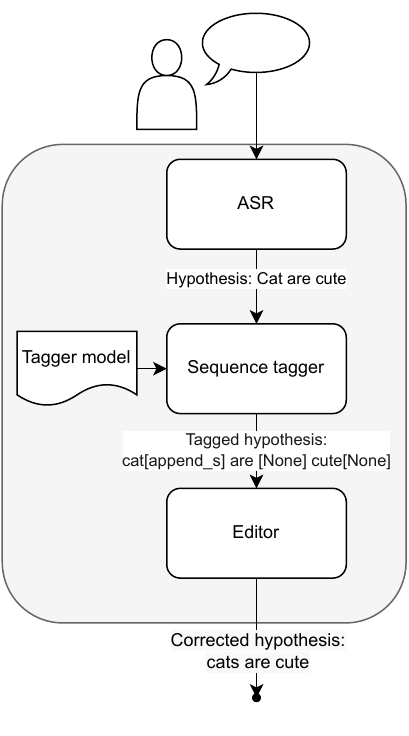}
     \caption{Error correction system architecture}
     \label{fig:runtime}

 \end{figure}

The proposed error correction method is designed to be precise, easily controllable, and data-efficient. In contrast to methods inspired by machine translation, such as \cite{guo_spelling_2019}, our model does not have to model and reproduce all tokens of the output sequence. Instead, it learns only which tokens to modify to correct the sentence.
To make it precisely controllable, the error correction mechanism works in two steps, depicted on figure \ref{fig:runtime}. First, a sequence-tagging model assigns a tag to each token in the input sentence. The tag indicates whether the token is correctly recognized or requires correction. In the latter case, the tag specifies an edit operation that is needed to transform the incorrect sentence into a correct one. In the second step, the assigned tags are used to perform edit operations that correct ASR errors. 
This approach is especially suitable for production settings because it allows one to precisely control which edit operations to include in the model and which edit operations to perform on inference time. The control can be based on scores returned by the tagger (for example, by setting a global scorer threshold) or on rules excluding certain operations in certain contexts from being performed.

\subsection{Tagging data preparation}\label{sec:tagging_data}
\begin{figure}[htbp]
  \centering
    \includegraphics[scale=1.0]{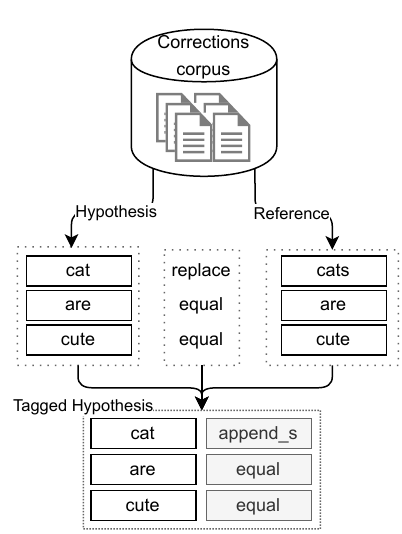}
    \caption{Tags generation example}
    \label{fig:example}
\end{figure}

To train a sequence tagging model, a corpus of ASR hypotheses with assigned edit operation tags is needed. We create it from the parallel corrections corpus described in Section \ref{sec:data}. First, hypothesis and reference sentences are compared and aligned using Ratcliff-Obershelp algorithm \cite{gestalt} implemented in difflib library \cite{difflib}. As a result, we get a list of operation codes describing how to turn corresponding parts of the hypothesis into reference. There are four operation codes: "replace", "delete", "insert" and "equal". For parts of a sentence which are not equal, we use a set of conditional rules involving recursively invoking the difflib alignment to tag each token with one of the edit operations. Examples of tags and corresponding edit operations are presented in table \ref{tab:operations}. For an illustrated example of tags generation process, see Figure \ref{fig:example}.

The set of available edit operation classes can be adjusted to the needs of a specific speech recognition system and a natural language. Ideally, the set should cover all errors and be as small as possible. If the set is too big, edit operations become too sparse for the tagger to learn them effectively. Therefore, we chose to prioritize use o 
most expressive edit operations, like ,,append\_s'', which for example in English could cover most of the errors associated with a singular noun in place of a plural and missing "s" in third-person singular verbs. The same errors could be covered by more precise rules correcting only particular words, e.g. "replace\_with\_cats" (edit operation assigned to the word "cat"). By using more general operations, the model can generalize to unseen examples of errors.
This is where our approach is different from \cite{malmi-etal-2019-encode} which uses only two main types of operations: KEEP ("None" in our approach), DELETE ("del" in our approach) and can insert any phrase or word from vocabulary $ V $ before the current token. Such an approach cannot cover multiple errors with one tag.

Despite using edit operations that cover multiple errors, the edit operations dataset is still sparse. About $15\%$ of edit operations are supported only with one example. To make training more efficient and the model less overfitted to singular examples, we use a cut-off of 150 most frequent edit operations, also filtering-out all which are found only once in the dataset. To differentiate between tokens which are correct and those, whose errors are not frequent enough, we replace all the filtered edit operations with a special edit operation called "unsupported". This operation tag is present in the training set and the model learns to tag some tokens with it, but when correcting sentences on the inference time, there is no edit operation performed on them. A sequence of edit operations assigned to adjacent tokens can be interdependent - performing only some of them may deteriorate the results. Therefore, when one of the tokens is tagged with "unsupported", all surrounding, non-empty tags are replaced with the "unsupported" tag.

\begin{table*}[ht]
    \caption{Examples of edit operations}
    \label{tab:operations}
    \centering
    \begin{tabular}{|l|l|l|}
      \hline
      \textbf{name} & \textbf{description} & \textbf{example}\\
      \hline
      \verb|del| & deletes a token & "a" $ \rightarrow $ "" \\
      \verb|append_s| & appends given suffix to the token & "cat" $ \rightarrow $ "cats" \\
      \verb|add_prefix_| & prepends given prefix to the token & "owl" $ \rightarrow $ "howl" \\
      \verb|remove_suffix_1| & removes 1 character from the end of the token & "cats" $ \rightarrow $ "cat" \\
      \verb|remove_prefix_1| & removes 1 character from the beginning of the token & "howl" $ \rightarrow $ "owl" \\
      \verb|join| & joins token with previous one & "book store" $ \rightarrow $ "bookstore" \\
       \verb|join_-| & joins token with previous one using given separator & "long term" $ \rightarrow $ "long-term" \\
      \verb|replace_| & replaces token with given string & "cat" $ \rightarrow $ "hat" \\
      \hline
    \end{tabular}
\end{table*}

\subsection{Tagging models}
We train and evaluate two tagging model types: BERT token classification model and contextual string embeddings model \cite{akbik-etal-2019-flair}, referred later as "Flair".
BERT tagger model uses locale variations of BERT transformer: Gbert for German \cite{gbert}, Camembert for French \cite{camembert} and Beto for Spanish \cite{beto-bert-spanish}. A single linear layer is added at the output, and the whole network is fine-tuned for the tagging task. 
Training was performed for 6 epochs, extending the training time did not improve results.
The "Flair" model contains Bert models mentioned before, extended with contextual string embeddings \cite{akbik-etal-2018-contextual}, LSTM \cite{lstm} and CRF \cite{Lafferty:2001:CRF} layers. Training was carried out for a maximum of 100 epochs.

\section{Results} \label{sec:results}
Both model types were evaluated using hold-out test sets by calculating WRR (Word Recognition Rate) of original ASR hypotheses and their corrected versions. Table \ref{tab:results} presents averaged results. As can be seen on an example of German dataset compared with other two - the worse the original ASR results are, the easier for the correction model to achieve higher absolute gains. Therefore, to compare correction models trained on datasets with different original results, we calculate a relative WER (Word Error Rate) reduction metric, which is calculated as 
\begin{equation}
    \frac{WER_{ASR} - WER_{Corrected}}{WER_{ASR}}
\end{equation} 
where $WER_{ASR}$ is WER of ASR system and $WER_{Corrected}$ is WER after applying the correction model. Relative improvements of our models vary between $21\%$ and $24.7\%$, making them comparable with state-of-the-art results reported in \cite{guo_spelling_2019}, while using much smaller training datasets, without the need of generating synthetic data with TTS engine.
The Flair tagger offers slightly better results (except for German, where results are equal).

	\begin{table}
		\caption{Results of error correction models}
        \label{tab:results}
        \centering
		\begin{tabular}{|l|l|r|r|r|}
    		\hline
    		& & \textbf{de-DE} & \textbf{es-ES} & \textbf{fr-FR} \\
    		\hline 
    		& base WRR & $78.07$ & $90.70$ & $93.51\%$   \\ 
    		\cline{2-5}
    		\multirow{3}{3em}{BERT} & corrected WRR & $83.48$ & $92.65$ & $94.97\%$  \\
    		& WRR gain & $5.41$ & $1.95$ & $1.46$  \\
    	    & Rel. WER reduction & $24.67\%$ & $20.97\%$ & $22.50\%$  \\
    	    \cline{2-5}
    	    \multirow{3}{3em}{Flair} & corrected WRR & $83.40$ & $92.86$ & $ 95.04 $  \\
    		& WRR gain & $5.33$ & $2.16$ & $1.53$  \\
    	    & Rel. WER reduction & $24.30\%$ & $23.23\%$ & $23.57\%$\\
    	    \hline
		\end{tabular}
    \end{table}

Table \ref{tab:times} presents the time required for training the models and the average times needed to correct a single sentence from the test set. Both training and inference were performed using a machine with a single Tesla P40 GPU.
	\begin{table}
		\caption{Training (in minutes) and inference (in milliseconds) times.}
        \label{tab:times}
        \centering
		\begin{tabular}{|l|l|r|r|r|}
    		\hline
    		& & \textbf{de-DE} & \textbf{es-ES} & \textbf{fr-FR} \\
    		\hline 
    		\multirow{3}{3em}{BERT} & training time  & $28$ & $40$ & $10$  \\
    		& inference latency  & $14$ & $14$ & $13$  \\
    	    \cline{2-5}
    	    \multirow{3}{3em}{Flair} & training time  & $180$ & $ 154 $ & $67$  \\
    		& inference latency  & $28$ & $29$ & $18$  \\
    	    \hline
		\end{tabular}
    \end{table}


\section{Conclusions} \label{sec:conclusions}
We presented a new approach to ASR errors correction problem. As demonstrated using three independent datasets, correction models trained using this approach are effective even for relatively small training datasets. The method allows to precisely control which errors should be included in the model and which of the included ones should be corrected at the inference time. The evaluations performed on the models show that they can significantly improve the ASR results by reducing the WER by more than $20\%$. All of the models presented offer very good inference latency, making them suitable for use with streaming ASR systems. 

The presented method is well suited for industrial applications where the ability to precisely control how the error correction model works, as well as small latency, are crucial. 


\bibliographystyle{IEEEtran}

\bibliography{mybib}

\end{document}